\documentclass[10pt,letterpaper]{article}

\usepackage{cogsci}

\cogscifinalcopy 

\usepackage[utf8]{inputenc}
\usepackage{pslatex}
\usepackage{caption}
\usepackage{graphicx}
\usepackage{subcaption} 
\usepackage{apacite}
\usepackage{amssymb}
\usepackage{amsmath}
\usepackage{amsfonts}   
\usepackage{float}

\newcommand{\bs}{\boldsymbol}
\newcommand{\x}{\boldsymbol{x}}

\newcommand{\thetab}{\boldsymbol{\theta}}

\newcommand{\given}{\,|\,}
\newcommand{\authorblock}[1]{\begin{tabular}{@{}c@{}}#1\end{tabular}}

\title{Amortized Bayesian Inference for Models of Cognition}
\author{\begin{tabular}{c@{\qquad}c}
  \authorblock{Stefan T.~Radev \\ \normalfont{Heidelberg University, Germany} \\ } &
  \authorblock{Andreas Voss \\ \normalfont{Heidelberg University, Germany} \\ } \\[\bigskipamount]
  \authorblock{Eva Marie Wieschen \\ \normalfont{Heidelberg University, Germany}\\ } &
  \authorblock{Paul-Christian Bürkner \\ \normalfont{Aalto University, Finland} \\ }
\end{tabular}}

\begin{document}

\maketitle

\begin{abstract}
As models of cognition grow in complexity and number of parameters, Bayesian inference with standard methods can become intractable, especially when the data-generating model is of unknown analytic form.
Recent advances in simulation-based inference using specialized neural network architectures circumvent many previous problems of approximate Bayesian computation.
Moreover, due to the properties of these special neural network estimators, the effort of training the networks via simulations amortizes over subsequent evaluations which can re-use the same network for multiple datasets and across multiple researchers. However, these methods have been largely underutilized in cognitive science and psychology so far, even though they are well suited for tackling a wide variety of modeling problems. With this work, we provide a general introduction to amortized Bayesian parameter estimation and model comparison and demonstrate the applicability of the proposed methods on a well-known class of intractable response-time models. 

\end{abstract}

\section{Generative Models in Cognitive Science}
Mathematical models formalize theories of cognition and enable the systematic investigation of cognitive processes through simulations and testable predictions. They enable a systematic joint analysis of behavioral and neural data, bridging a crucial gap between cognitive science and neuroscience \cite{turner2019joint}. Moreover, questions demanding a choice among competing cognitive theories can be resolved at the level of formal model comparison.  

The \textit{generative} property of such models arises from the fact that one can simulate the process of interest and study how it behaves under various conditions. More formally, consider a cognitive model $\mathcal{M}$ which represents a theoretically plausible, potentially noisy, process by which observable behavior $x$ arises from an assumed cognitive system governed by hidden parameters $\theta$ and an independent source of noise $\xi \sim p(\xi)$:
\begin{equation}
     x = \mathcal{M}(\theta, \xi)  \label{eq:1}
\end{equation}
Generative models of this form have been developed in various domains throughout psychology and cognitive science, including decision making \cite{voss2019sequential}, memory \cite{myung2007analytic}, reinforcement learning \cite{fontanesi2019reinforcement}, risky behavior \cite{stout2004cognitive}, to name just a few. Once a model (or a set of models) of some cognitive process of interest has been formulated, the challenge becomes to perform inference on real data. We will now briefly review the mathematical tools provided by Bayesian probability theory for parameter estimation and model comparison \cite{jaynes2003probability}. Then, we will peruse a novel framework for performing Bayesian inference on models of cognition which are intractable with standard Bayesian methods.

\section{Bayesian Parameter Estimation}

Bayesian parameter estimation leverages prior knowledge about reasonable parameter ranges and integrates this information with the information provided by the data to arrive at a \textit{posterior distribution} over parameters. In a Bayesian context, the posterior encodes our updated belief about plausible parameter ranges conditional on a set of $N$ observations $X := \{x_n\}_{n=1}^N$. Bayes' rule gives us the well known analytical form of the posterior:
\begin{equation}
    p(\theta \given X) = \frac{p(X \given \theta)\,p(\theta)}{\int p(X \given \theta)\,p(\theta)\,d \theta} \label{eq:2}
\end{equation}
where $p(X \given \theta)$ represents the \textit{likelihood} of the parameters $\theta$ and $p(\theta)$ denotes the \textit{prior}, that is the distribution of $\theta$ before observing the data. The denominator is a normalizing constant usually referred to as the \textit{marginal likelihood} or \textit{evidence}. Note, that all distributions are also implicitly conditional on the particular generative model $\mathcal{M}$.

Based on the obtained estimate of the posterior distribution, usually in the form of random draws from the posterior, summary statistics such as posterior means or credible intervals for each parameter can be obtained.
What is more, the posterior distribution can be further transformed to obtain subsequent quantities of interest, for example, the \textit{posterior predictive distribution} which can be compared to the observed data for the purpose of model checking \cite{lynch2004bayesian}.



\section{Bayesian Model Comparison}

In many research domains, there is not a \textit{single model} for a particular process, but whole \textit{classes} of models instantiating different and often competing theories. Bayesian model comparison proceeds by assigning a plausibility value to each candidate model. These plausibility values (model weights, model probabilities, model predictions, etc.) can be used to guide subsequent model selection. 

To set the stage, consider a set of $J$ candidate models $\mathcal{G} = \{\mathcal{M}_1, \mathcal{M}_2,\dots,\mathcal{M}_J\}$. An intuitive way to quantify plausibility is to consider the marginal likelihood of a model $\mathcal{M}$ given by:
\begin{equation}
     p(X \given \mathcal{M}) = \int p(X \given \theta, \mathcal{M})\,p(\theta \given \mathcal{M})\,d\theta
\end{equation}
which is also the denominator in Eq.\ref{eq:2} (with $\mathcal{M}$ implicit in the previous definition). This quantity is also known as \textit{evidence}, or \textit{prior predictive} distribution, since the likelihood is weighted by the prior (in contrast to a posterior predictive distribution where the likelihood would be weighted by the posterior). The marginal likelihood penalizes the prior complexity of a model and thus naturally embodies the principle of Occam's razor \cite{jaynes2003probability}. To compare two competing models, one can focus on the ratio between two marginal likelihoods, called a Bayes factor (BF):
\begin{equation}
    \textrm{BF}_{ij} = \frac{p(X \given \mathcal{M}_{i})}{p(X \given \mathcal{M}_j)}
\end{equation}
which quantifies the relative evidence of model $i$ over model $j$. Alternatively, if prior information about model plausibility is available, one can consider model posteriors $p(\mathcal{M} \given X) \propto p(X \given \mathcal{M})\,p(\mathcal{M})$ and compute the posterior odds:
\begin{equation}
    \frac{p(\mathcal{M}_i \given X)}{p(\mathcal{M}_j \given X)} = \frac{p(X \given \mathcal{M}_{i})}{p(X \given \mathcal{M}_j)}\,\frac{p(\mathcal{M}_i)}{p(\mathcal{M}_j)}
\end{equation}
which combine the relative evidence given by the BF with prior information in the form of prior odds. 

\section{Model Intractability}

In order for cognitive models to be useful in practice, parameter estimation and model comparison should be feasible within reasonable time limits. As evident from their definitions, both Bayesian parameter estimation and model comparison depend on the likelihood function $p(X \given \theta, \mathcal{M})$ which needs to be evaluated analytically or numerically for any triplet $(\mathcal{M}, \theta, X)$. 

When this is possible, standard Bayesian approaches for obtaining random draws from the posterior, such as Markov chain Monte Carlo (MCMC), or optimizing an approximate posterior, such as variational inference (VI), can be readily applied. However, when the likelihood function is not available in closed-form or too expensive to evaluate, standard methods no longer apply.

In fact, many interesting models from a variety of domains in cognitive science and psychology turn out to be intractable \cite{voss2019sequential, turner2016bayesian}. This has precluded the wide exploration and application of these models, as researchers have often traded off complexity or neurocognitive plausibility for simplicity in order to make these models tractable. In the following, we discuss the most popular approach to inference with intractable models.

\section{Simulation-Based Inference}

Simulation-based methods leverage the generative property of mathematical models by treating a particular model as a \textit{scientific simulator} from which synthetic data can be obtained given any configuration of the parameters. Simulation-based inference is common to many domains in science in general \cite{cranmer2019frontier} and a variety of different approaches exist. These methods have also been dubbed \textit{likelihood-free}, which is somewhat unfortunate, since the likelihood is implicitly defined by the generative process and sampling from the likelihood is realized through the stochastic simulator:
\begin{equation}
     \x_n \sim p(\x \given \thetab, \mathcal{M}) \Longleftrightarrow\ x_n = \mathcal{M}(\thetab,\xi_n) \textrm{ with } \bs{\xi}_n \sim p(\bs{\xi}) 
\end{equation}
Different simulation-based methods differ mainly with respect to how they utilize the synthetic data to perform inference on real observed data \cite{cranmer2019frontier}. The utility of any simulation-based method depends on multiple factors, such as asymptotic guarantees, data utilization, efficiency, scalability, and software availability.

Approximate Bayesian computation (ABC) offers a standard set of theoretically sound methods for performing inference on intractable models \cite{cranmer2019frontier}. The core idea of ABC methods is to approximate the posterior by repeatedly sampling parameters from a proposal (prior) distribution and then generating a synthetic dataset by running the simulator with the sampled parameters.
If the simulated dataset is sufficiently similar to an actually observed dataset, the corresponding parameters are retained as a sample from the desired posterior, otherwise rejected. However, in practice, ABC methods are notoriously inefficient and suffer from various problems, such as the \textit{curse of dimensionality} or \textit{curse of inefficiency} \cite{marin2018likelihood}. More efficient methods employ various techniques to optimize sampling or correct potential biases. 

Recently, the scientific repertoire for simulation-based inference has been enhanced with ideas from deep learning and neural density estimation (NDE) in particular \cite{greenberg2019automatic}. 
These methods employ specialized neural network architectures which are trained with simulated data to perform efficient and accurate inference on previously intractable problems \cite{cranmer2019frontier}. 
NDE methods are rapidly developing and still largely underutilized in cognitive modeling, even though first applications to simulated \cite{radev2020bayesflow, radev2020amortized} as well as actual data \cite{wieschen2020jumping} exist.

\section{Amortized Inference}

\begin{figure*}
\centering
\begin{subfigure}{.99\textwidth}
    \includegraphics[width=\textwidth]{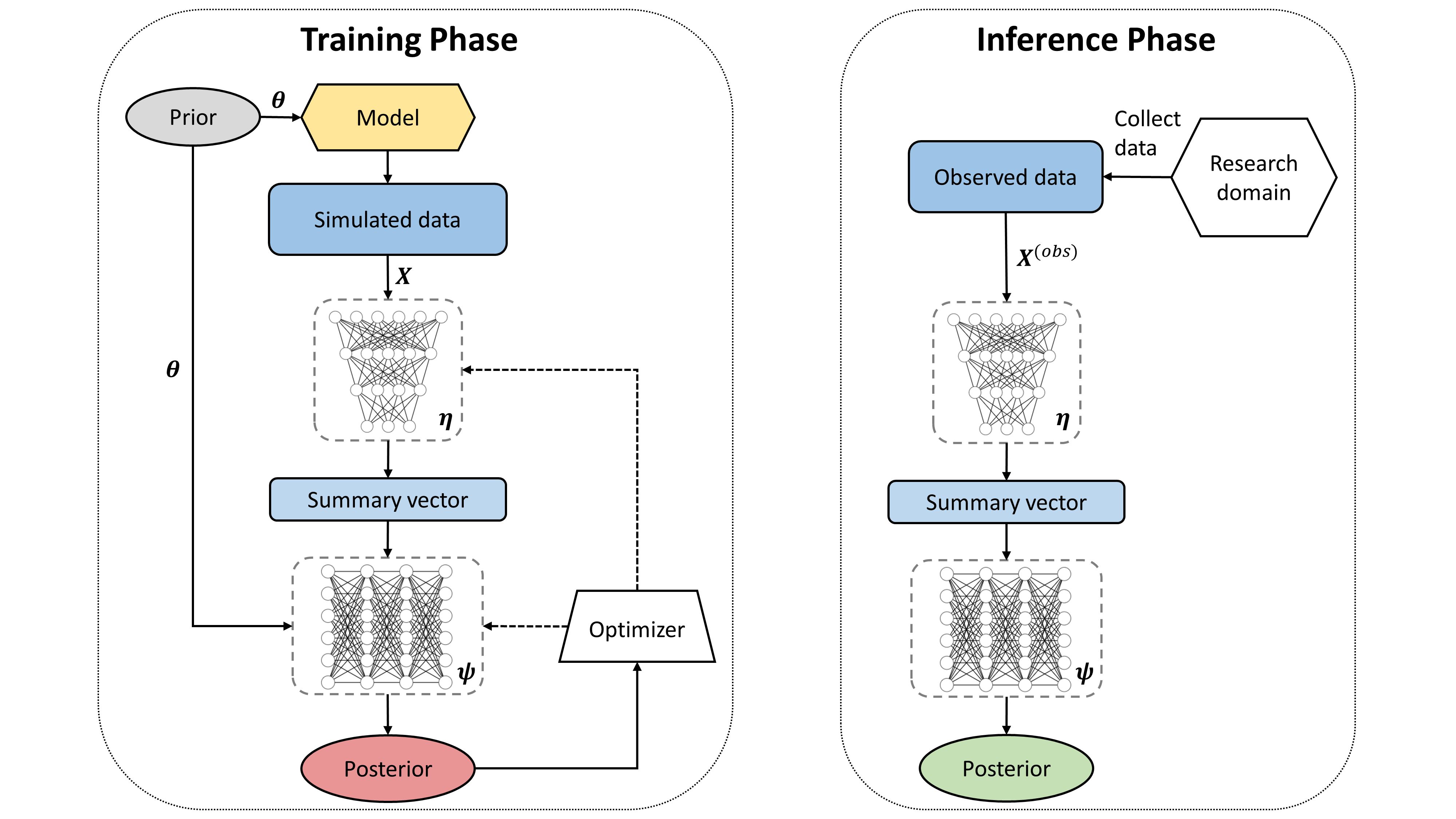}
    \caption{Amortized parameter estimation}
    \label{fig:Fig.1a}
\end{subfigure}
\begin{subfigure}{.99\textwidth}
    \includegraphics[width=\textwidth]{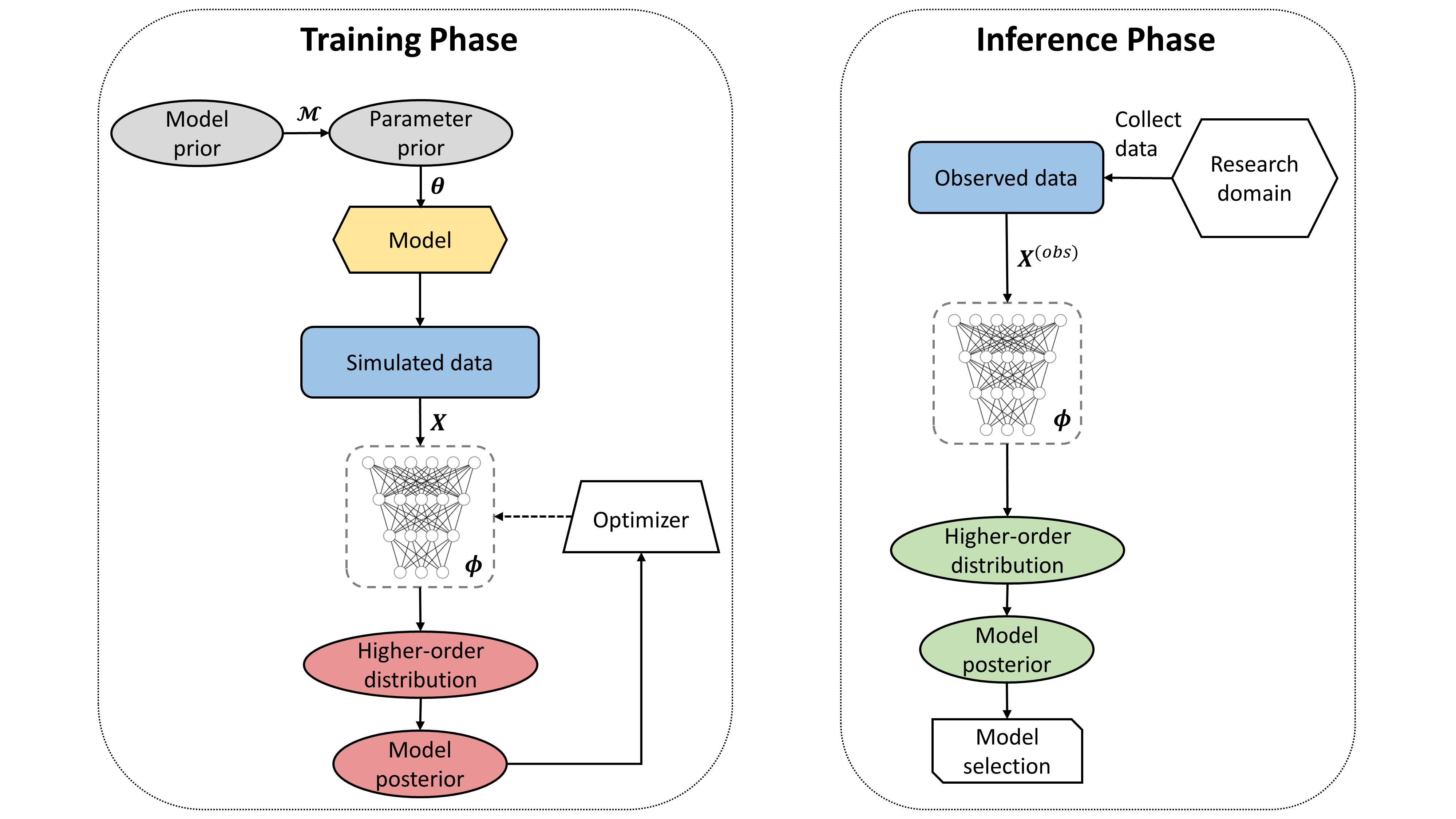}
    \caption{Amortized model comparison}
    \label{fig:Fig.1b}
\end{subfigure}
\caption[short]{Graphical illustration of amortized parameter estimation and model comparison with different neural network estimators. (\textbf{a}) Amortized Bayesian parameter estimation with invertible neural networks \cite{radev2020bayesflow}. The left panel depicts the training phase in which the summary ($f_{\eta}$) and the inference network ($f_{\psi}$) are jointly optimized to approximate the true target posterior. The right panel depicts inference with already trained networks on observed data; (\textbf{b}) Amortized Bayesian model comparison with evidential neural networks \cite{radev2020amortized}. The left panel depicts the training phase during which the evidential network $f_{\phi}$ is optimized to approximate the true model posteriors via a higher-order Dirichlet distribution. The right panel depicts inference with an already trained evidential network; the upfront training effort for both inference tasks is amortized over arbitrary numbers of datasets from a research domain.}
\label{fig:Fig.1}
\end{figure*}

The majority of simulation-based methods need to be applied to each dataset separately. This quickly becomes infeasible when multiple datasets are to be analysed and multiple candidate models are considered, since the expensive inference procedure needs to be repeated from scratch for each combination of dataset and model.  

In contrast, the concept of \textit{amortized inference} refers to an approach which minimizes the cost of inference by separating the process into an expensive training (optimization) phase and a cheap inference phase which can be easily repeated for multiple datasets or models without computational overhead. Thus, the effort of training or optimization amortizes over repeated applications on multiple datasets or models. In some cases, the efficiency advantage of amortized inference becomes noticeable even for a few datasets \cite{radev2020bayesflow, radev2020amortized}. 

The field of amortized inference is rapidly growing and a variety of methods and concepts are currently being explored. 
For instance, \textit{inference compilation} involves pre-training a neural network with simulations from a generative model and then using the network in combination with a probabilistic program to optimize sampling from the posterior \cite{le2016inference}.
The \textit{pre-paid} estimation method \cite{mestdagh2019prepaid} proceeds by creating a large grid of simulations which are reduced to summary statistics and stored on disk. Subsequent inference involves computing the nearest neighbors of an observed dataset in the pre-paid grid and interpolation.
Sequential neural posterior estimation (SNPE) methods employ various iterative refinement schemes to transform a proposal distribution into the correct target posterior via expressive NDEs trained over multiple simulation rounds \cite{greenberg2019automatic}.

In line with these ideas, we recently proposed two general frameworks for amortized Bayesian parameter estimation and model comparison based on specialized neural network architectures \cite{radev2020bayesflow, radev2020amortized}. In particular, these frameworks were designed to implement the following desirable properties: 
\begin{itemize}
    \setlength\itemsep{0.2em}
    \item Fully amortized Bayesian inference for parameter estimation and model comparison of intractable models
    \item Asymptotic theoretical guarantees for sampling from the true parameter and model posteriors
    \item Learning maximally informative summary statistics directly from data instead of manual selection 
    \item Scalability to high-dimensional problems through considerations regarding the probabilistic symmetry of the data
    \item Implicit preference for simpler models based purely on generative performance
    \item Online learning eliminating the need for storing large grids or reference tables
    \item Parallel computations and GPU acceleration applicable to both simulations, training/optimization, and inference
\end{itemize}

In the following, we describe our recently developed methods parameter estimation and model comparison in turn. 

\section{Amortized Parameter Estimation with Invertible Neural Networks}

Recently, we proposed a novel amortization method based on invertible neural networks \cite{radev2020bayesflow}, which we dubbed \textit{BayesFlow}. The method relies solely on simulations from a process model in order to learn and calibrate the full posterior over all possible parameter values and observed data patterns.

The BayesFlow method involves two separate neural networks trained jointly. A permutation invariant \textit{summary network} is responsible for reducing an entire dataset $X$ with a variable number $N$ of $i.i.d.$ observations\footnote{Note, that the \textit{i.i.d.} assumption is not a necessary condition for the method to work, but used here only to simplify the discussion.} into a vector of \textit{learned summary statistics}. Importantly, permutation invariant networks can deal with \textit{i.i.d.} sequences of variable size and preserve their probabilistic symmetry. An \textit{inference network}, implemented as an invertible neural network \cite{radev2020bayesflow}, is responsible for approximating the true posterior of model parameters given the output of the summary network. 
Invertible networks can perform asymptotically exact inference and scale well from simple low-dimensional problems to high-dimensional distributions with complex dependencies. 
During training, model parameters and synthetic datasets are generated on the fly and neural network parameters are adjusted via joint backpropagation (see Figure \ref{fig:Fig.1a}, left panel, for a graphical illustration of the training phase).

Given a model and a prior over the model parameters, the goal is thus to train a conditional invertible neural network $f_{\psi}$ with adjustable parameters $\psi$ together with a summary network $f_{\eta}$ with adjustable parameters $\eta$. These networks jointly learn an approximate posterior $p_{\psi}(\theta \given f_{\eta}(X))$ over the relevant parameters for arbitrary numbers of datasets and dataset sizes $N$, as long as they share the same data structure. To achieve this, the networks minimize the Kullback-Leibler (KL) divergence between the true and the approximate posterior:
\begin{align}
   \min_{\psi, \eta} \mathbb{KL} \left(p(\theta \given X)\,||\, p_{\psi}(\theta \given f_{\eta}(X))\right) 
\end{align}
\noindent Utilizing the fact that we have access to the joint distribution $p(\thetab, X) = p(\theta)\,(X \given \theta)$ via the simulator, we minimize the KL divergence in expectation over all possible datasets that can be generated given the prior and the model, resulting in the following optimization criterion:
\begin{equation}
\min_{\psi, \eta} \mathbb{E}_{p(\theta, x)} \left[  -\log p_{\psi}(\theta \given f_{\eta}(X)) \right]
\end{equation}
\noindent In practice, we approximate the criterion via its Monte Carlo (MC) estimate, since we can simulate theoretically infinite amounts of data and can easily evaluate $p_{\psi}(\theta \given f_{\eta}(X))$ due to our invertible architecture. 
 In case of perfect convergence of the networks, the summary network outputs \textit{sufficient summary statistics} and the inference network samples from the true posterior \cite{radev2020bayesflow}.
 Importantly, once the networks have been trained with sufficient amounts of simulated data, they can be stored and applied for inference on multiple datasets from a research domain (see Figure \ref{fig:Fig.1a}, right panel).

\section{Amortized Model Comparison with Evidential Neural Networks}

In another recent work \cite{radev2020amortized}, we explored a framework for Bayesian model comparison on intractable models via \textit{evidential neural networks}. We proposed to train a permutation invariant classifier network on simulated data from multiple models. The goal of this network is to approximate posterior model probabilities as accurately as possible. To achieve this, the network is trained to output the parameters of a higher-order probability distribution (parameterized as a Dirichlet distribution) over the model probabilities themselves, which quantifies the uncertainty in model probability estimates. Thus, for a classifier network with parameters $\phi$, the higher-order posterior distribution over model probabilities is given by:
\begin{equation}
    \textrm{Dir}(\pi \given \alpha_{\phi}(X)) = \frac{1}{B(\alpha_{\phi}(X))} \prod_{j=1}^J\, \pi^{\alpha_{\phi}(X)_j - 1}
\end{equation}
where $\alpha_{\phi}(X)$ denotes the vector of \textit{concentration parameters} obtained by the network for a dataset $X$ and $B(\cdot)$ is the multivariate \textit{beta} function. The mean of this Dirichlet distribution can be used as a best estimate for the posterior model probabilities:
\begin{equation}
    p_{\phi}(\mathcal{M} \given X) = \frac{\alpha_{\phi}(X)}{\sum_{j=1}^J \alpha_{\phi}(X)_j}
\end{equation}
Additionally, its variance can be interpreted as the epistemic uncertainty surrounding the actual evidence which the data provide for model comparison.

For training the network, we again utilize the fact that we have access to the joint distribution $p(\mathcal{M}, \theta, X)$ via simulations (see Figure \ref{fig:Fig.1b}, left panel). Our optimization criterion is:
\begin{equation}
    \min_{\phi}\mathbb{E}_{p(\mathcal{M},\theta, X)} \left[\mathcal{L} \left( p_{\phi}(\mathcal{M} \given X), \mathcal{M} \right) \right]
\end{equation}
where $\mathcal{L}(\cdot,\cdot)$ is a \textit{strictly proper} loss function \cite{gneiting2007strictly}, $\mathcal{M}$ is the true model index and the data $X$ implicitly depend on $\theta$. In practice, we approximate this expectation via draws from the joint distribution available via the simulator. Optimization of a strictly proper criterion, asymptotic convergence implies that the mean of the Dirichlet distribution represents the true model posteriors. Moreover, our simulation-based approach implicitly captures a preference for simpler models (Occam's razor), since simpler models will tend to generate more similar datasets. As a consequence, when such datasets are plausible under multiple models, the comparably simpler models will be more probable.  

As with parameter estimation, once the evidence network has been trained on simulated data from the candidate models, it can be applied to multiple upcoming observations from a research domain (see Figure \ref{fig:Fig.1b}, right panel).

\section{Example Applications}

In the following, we will present two applications of amortized Bayesian parameter estimation to a recently proposed and intractable evidence accumulation model (EAM). The first illustrative application is a simulation study aimed at quantifying parameter recovery as a function of data set size. Such simulations are especially useful for planing experiments but usually too costly to perform in complex modeling scenarios.
The second application is concerned with parameter estimation on real data and serves as an illustration on how researchers might utilize amortized Bayesian inference with a pre-trained density estimator in practice.

EAMs are a popular class of models in psychology and cognitive science, as they allow a model-based analysis of response time (RT) distributions. 
Here, we will consider a Lévy flight model (LFM) with a non-Gaussian noise assumption \cite{voss2019sequential, wieschen2020jumping} as an example. The Lévy flight process is driven by the following stochastic ordinary differential equation (ODE):
\begin{align}
dx_c &= v_c\,dt + \xi dt^{1/\alpha}  \\
\xi &\sim \textrm{AlphaStable}(\alpha,0,1,0) 
\end{align}
where $dx_c$ denotes accumulated cognitive evidence in condition $c$, $v_c$ denotes the average speed of information accumulation (drift), and $\alpha$ controls how heavy the tails of the noise distribution are (i.e., smaller values increase the probability of outliers in the accumulation process). Further parameters of the model are: a decision threshold ($a$) which reflects the amount of information needed for selecting a response; a starting point ($z_r$) indicative of response biases; and a non-decision time ($t_0$) reflecting additive encoding and motor process. 
Since the relationship of the $\alpha$ parameter to the standard parameters of the classical diffusion model \cite{ratcliff2004diffusion} has not been previously investigated, we focus on quantifying posterior correlations in the real data application. 

\subsection{Simulation Example}

\begin{figure}
\centering
\begin{subfigure}{.49\textwidth}
    \includegraphics[width=\textwidth]{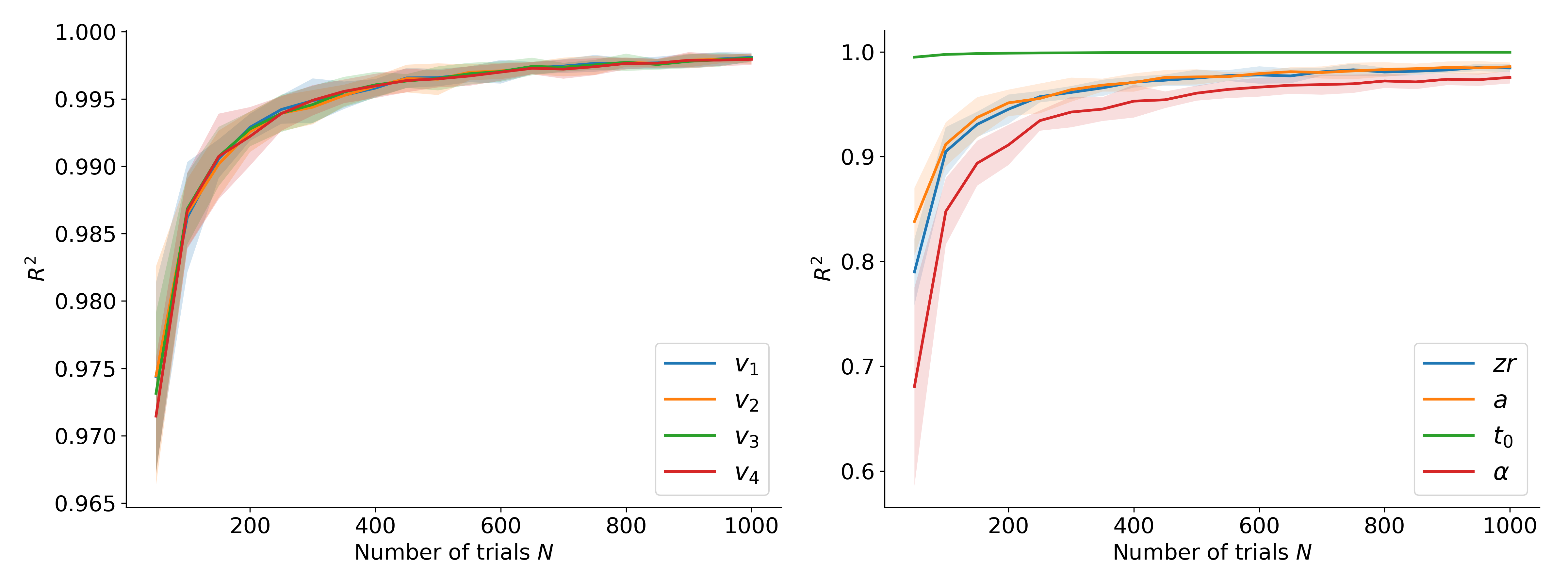}
    \caption{Parameter recovery as a function of trial numbers ($N$)}
    \label{fig:Fig.2a}
\end{subfigure}
\begin{subfigure}{.49\textwidth}
    \includegraphics[width=\textwidth]{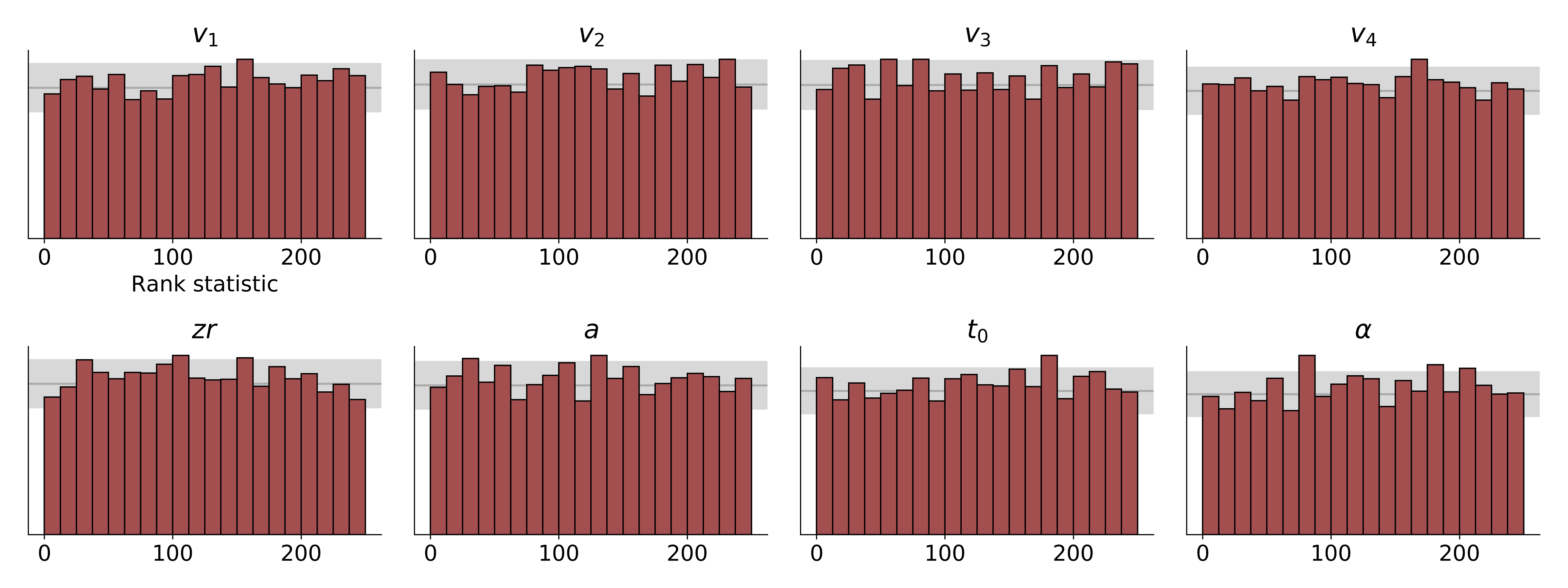}
    \caption{Simulation-based calibration}
    \label{fig:Fig.2b}
\end{subfigure}
\caption[short]{Simulation results. (\textbf{a}) The left panel depicts parameter recovery of the four drift rate parameters as a function of trial numbers per participant $N$. The right panel depicts recovery of the other four parameters. Posterior means are used as summaries of the full posteriors and shaded regions represent bootstrap 95\% confidence intervals. (\textbf{b}) The panel depicts simulation-based calibration (SBC) results at $N=800$ as a validation check for the correctness of the full posteriors.}
\label{fig:Fig.2}
\end{figure}

As a first example, consider a simulated RT experiment with four conditions. How many trials are needed for accurate parameter recovery? To answer this question, we can simulate multiple experiments with varying number of trials per participant ($N$) and then compute some discrepancy between ground-truth parameters and their estimates.
\begin{figure}
\centering
\begin{subfigure}{.49\textwidth}
    \includegraphics[width=\textwidth]{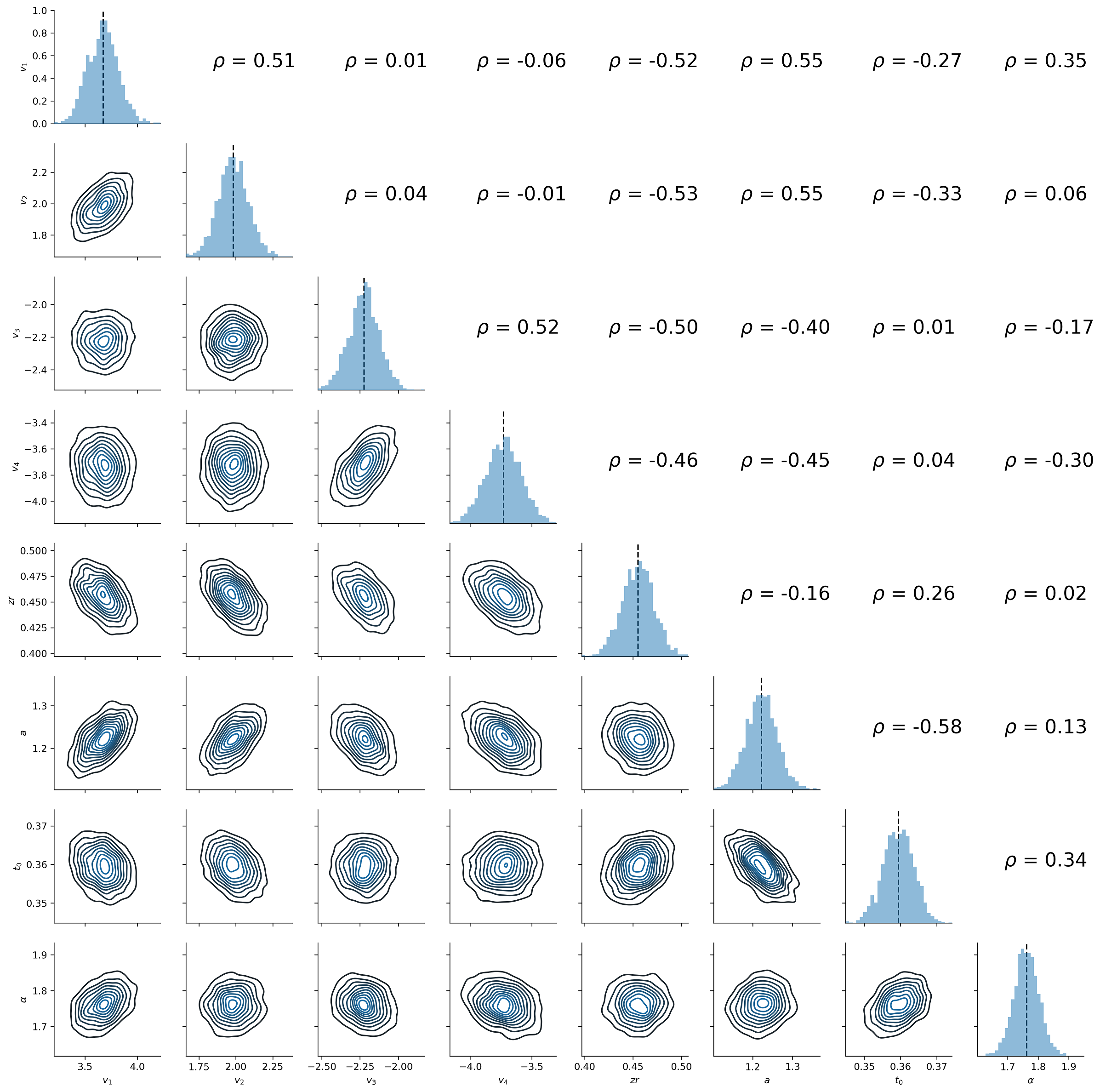}
\end{subfigure}
\caption[short]{Example full posteriors and bivariate posterior correlations obtained from data of one participant in the long LDT via amortized Bayesian inference. Dashed lines on the main diagonal indicate posterior means.}
\label{fig:Fig.3}
\end{figure}
However, since the model is intractable, such a simulation scenario is not feasible with non-amortized methods, which would need weeks on standard machines \cite{voss2019sequential}. However, using the \textit{BayesFlow} method (Figure \ref{fig:Fig.1a}), we can train the networks with simulated datasets and vary the number of trials during each simulation. Such a training takes approximately one day on a standard laptop equipped with an NVIDIA\textsuperscript{\textregistered} GTX1060 graphics card. Subsequent inference is then very cheap, as amortized parameter estimation on 500 simulated participants takes less than 2 seconds. 

We visualize the results by plotting the average $R^2$ metric obtained from fitting the LFM model to $300$ simulated participants at different $N$ between $50$ and $1000$ (see Figure \ref{fig:Fig.2a}). Notably, recovery of the ground-truth parameters via posterior means is nearly perfect at higher trial numbers.

As a validation tool for visually detecting systematic biases in the approximate posteriors, we can also cheaply apply simulation-based calibration (SBC) and inspect the rank statistic of the posterior samples for uniformity \cite{talts2018validating}. Results from applying SBC to $5000$ simulated participants at $N=800$ are depicted in Figure \ref{fig:Fig.2b}. Indeed, we confirm that no pronounced issues across all marginal posteriors are present.

\subsection{Real Data Example}

We can also apply the same networks from the previous simulation example for fully Bayesian inference on real data. Here, we fit the LFM model to previously unpublished data from eleven participants performing a long ($N=800$ per condition) lexical decision task (LDT). Since the task had a $2 \times 2$ design, with a factor for \textit{difficulty} (hard vs. easy), and a factor for \textit{stimulus type} (word vs. non-word), we can assume a different drift rate for each design cell.

Applying the pre-trained networks, we immediately obtain samples from a full posterior over model parameters for each participant. 
Using the estimated posteriors, we can then test hypotheses about particular parameter values, compute individual differences, or compare means between conditions in a Bayesian way. 
Furthermore, we can analyze posterior correlations at an individual level and investigate task-dependent relationships between the $\alpha$ parameter and other parameters (see Figure \ref{fig:Fig.3} for results obtained from a single participant). 

Across participants, $\alpha$ displays only small posterior correlations with drift rates as well as small posterior correlations with threshold and non-decision time parameters (mean $r < 0.5$ across all parameters of the standard diffusion model). 
These results provide first evidence that the $\alpha$ parameter can indeed be decoupled from other model parameters and possibly indicates a separate decision process. 

Since the goal of this application was solely to illustrate a typical use case for amortized Bayesian inference, future research should focus on extensive external validation of the LFM model as well as proposing a neurocognitively plausible interpretation for the $\alpha$ parameter.  

\section{Outlook}

The purpose of this work was to introduce the main ideas behind amortized Bayesian inference methods for simulation-based parameter estimation and model comparison. Although these methods come with promising theoretical guarantees and clear practical advantages, their utility for cognitive modeling is just beginning to be explored. Moreover, there are still many open questions and avenues for future research.

First, a systematic investigation of a potential \textit{amortization gap} in certain practical application seems warranted. An amortization gap refers to a drop in estimation accuracy due to the fact that we are relying on a single set of neural network parameters for solving an inference problem globally, instead of performing per-dataset optimization. Even though we have not observed such a scenario in our applications and simulations, this behavior might occur when the neural network estimators are not expressive enough to represent complex posterior distributions. 

Second, there are still little systematic guidelines on how to best design and tune the neural network architectures so as to perform optimally across a variety of parameter estimation and model comparison tasks. Even though neural density estimation methods outperform standard ABC methods on multiple metrics and in various contexts, there is certainly room for improvement. Black-box optimization methods for hyperparameter tuning, such as Bayesian optimization or active inference \cite{snoek2012practical}, might facilitate additional performance gains and reduce potentially suboptimal architectural choices. 

Finally, user-friendly software for applying Bayesian amortization methods \textit{out-of-the-box} is still largely in its infancy. Developing and maintaining such software is a crucial future goal for increasing the applicability and usability of novel simulation-based methods.  

\section{Conclusion}

We hope that the inference architectures discussed in this work will spur the interest of cognitive modelers from various domains. We believe that such architectures can greatly enhance model-based analysis in cognitive science and psychology. By leaving subsidiary tractability considerations to powerful end-to-end algorithms, researchers can focus more on the task of model development and evaluation to further improve our understanding of cognitive processes.

\section{Acknowledgments}

This research was supported by the Deutsche Forschungsgemeinschaft (DFG, German Research Foundation; grant number GRK 2277 "Statistical Modeling in Psychology"). We thank the Technology Industries of Finland Centennial Foundation (grant 70007503; Artificial Intelligence for Research and Development) for partial support of this work.

\bibliographystyle{apacite}
\setlength{\bibleftmargin}{.125in}
\setlength{\bibindent}{-\bibleftmargin}
\bibliography{references}

\end{document}